\documentclass[letterpaper, 10pt, conference]{ieeeconf}

\IEEEoverridecommandlockouts
\usepackage{graphicx}
\usepackage{amsmath}
\usepackage{amssymb}
\usepackage{mathnotation}
\usepackage{cite}
\usepackage{mathtools}
\usepackage[hidelinks]{hyperref}

\begin{document}

\title{Geometric Design and Gait Co-Optimization for Soft Continuum Robots Swimming at Low and High Reynolds Numbers}

\author{Yanhao Yang, Ross L. Hatton 
\thanks{This work was supported in part by NSF Grants No. 1653220, 1826446, and 1935324, and by ONR Grant No. N00014-23-1-2171.}
\thanks{Y. Yang and R. L. Hatton are with the  Collaborative Robotics and Intelligent Systems (CoRIS) Institute at Oregon State University, Corvallis, OR USA. {\tt\small \{yangyanh, Ross.Hatton\}@oregonstate.edu}} 
}

\maketitle

\begin{abstract}

Recent advancements in soft actuators have enabled soft continuum swimming robots to achieve higher efficiency and more closely mimic the behaviors of real marine animals. However, optimizing the design and control of these soft continuum robots remains a significant challenge. In this paper, we present a practical framework for the co-optimization of the design and control of soft continuum robots, approached from a geometric locomotion analysis perspective. This framework is based on the principles of geometric mechanics, accounting for swimming at both low and high Reynolds numbers. By generalizing geometric principles to continuum bodies, we achieve efficient geometric variational co-optimization of designs and gaits across different power consumption metrics and swimming environments. The resulting optimal designs and gaits exhibit greater efficiencies at both low and high Reynolds numbers compared to three-link or serpenoid swimmers with the same degrees of freedom, approaching or even surpassing the efficiencies of infinitely flexible swimmers and those with higher degrees of freedom.

\end{abstract}

\section{Introduction}

Swimming has attracted significant attention across various fields, including fluid mechanics \cite{taylor1951analysis, wu1961swimming, cox1970motion, lighthill1971largeamplitude, purcell1977life, lauder2007fish, liao2007review, candelier2011threedimensional}, biology \cite{blake1983fish, alexander2006principles}, and robotics \cite{mcisaac2003motion, colgate2004mechanics, morgansen2007geometric, hirose2009snakelike}. The study of the mechanisms behind the undulatory self-propulsion of aquatic organisms has provided valuable insights \cite{kanso2005locomotion, kanso2009swimming, boyer2010poincare, candelier2011threedimensional}, inspiring a wide range of research efforts focused on the design and control of robotic swimmers \cite{melli2006motion, boyer2008fast, boyer2012macrocontinuous, hatton2013geometric, porez2014improved}.

Swimming robots are not always driven by discrete joint motors. Recent advances in soft actuators and artificial muscles are driving the development of swimming robots with soft continuum designs \cite{burdick1994sidewinding, boyer2012macrocontinuous, marchese2014autonomous, burgner-kahrs2015continuum, rus2015design, aguilar2016review, blumenschein2018helical, rich2018untethered, kim2019ferromagnetic, schiebel2020robophysical, liu2021design, chong2022coordinating, gallentine2022multimodal, vanstratum2022comparative}, which offer higher efficiency, compliance, and safety, as well as more closely resemble natural swimming organisms. These soft continuum swimming robots introduce new challenges in design, modeling, and motion planning: What is the optimal shape for artificial muscles for swimming? How does design influence motion planning?

Geometric mechanics can reduce the complex, high-dimensional problems found in robotics to simpler, low-dimensional ones, for design, analysis, and planning \cite{melli2006motion, shammas2007geometric, hatton2013geometric, ramasamy2019geometry, hatton2022geometry, chong2022coordinating, yang2024geometric}. This approach has become a powerful tool for studying swimming due to its efficiency, interpretability, and consistency across various types of locomotion and environments. Geometric mechanics methods have been applied to the design \cite{chong2022coordinating, sparks2022amoebainspired, rozaidi2023hissbot}, system identification \cite{bittner2018geometrically}, and motion planning \cite{ramasamy2019geometry, hatton2022geometry, yang2024geometric} of swimming robots across different Reynolds numbers (Re). While there have been attempts to apply these principles to the design optimization of swimmers with variable link lengths \cite{ramasamy2019geometry}, there has been no dedicated research on applying geometric mechanics to the co-optimization of the design and control of soft continuum robots.

\begin{figure}[!t]
\centering
\includegraphics[width=\linewidth]{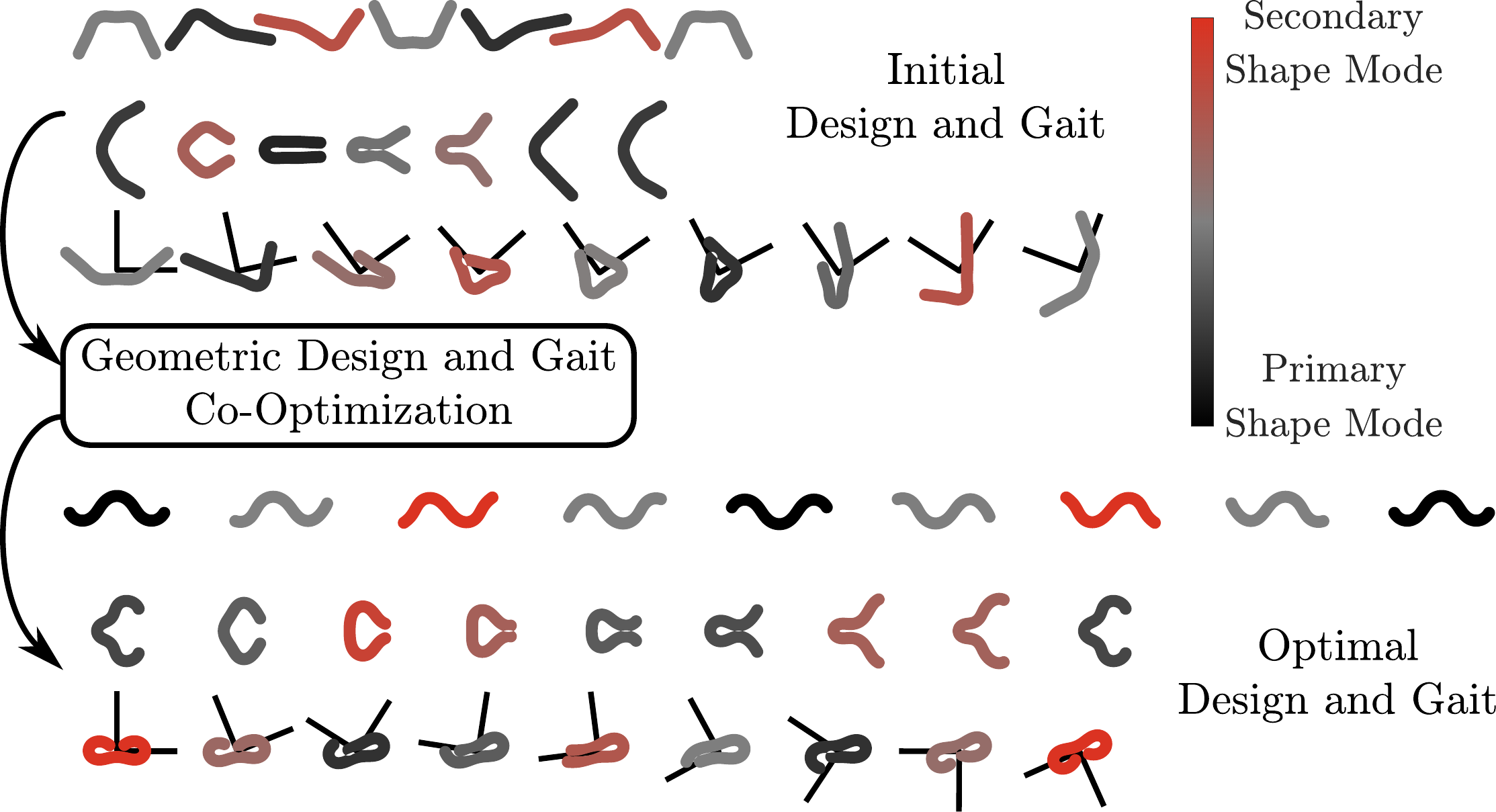}
\caption{Schematic diagram of the co-optimization of design and control for soft continuum robots based on geometric mechanics. Starting with an initial guess for the design and gait of the soft continuum robot, we apply geometric mechanics principles to model continuum swimming to obtain the gradients of displacement and power consumption relative to the design and gait. These gradients are then used in variational optimization to determine the optimal design and gait for the soft continuum swimming robot.}
\label{fig:introduction}
\end{figure}

In this paper, we introduce a practical framework for the co-optimization of the design and control of soft continuum robots, utilizing the geometric locomotion analysis approach, as illustrated in Fig.~\ref{fig:introduction}. This framework leverages geometric mechanics principles to analyze swimming at low and high Re. By extending geometric principles to continuum bodies, we enable geometric variational co-optimization of designs and gaits across various power consumption metrics and swimming environments. The resulting optimal designs and gaits demonstrate superior efficiency at both low and high Re compared to three-link or serpenoid swimmers with the same degrees of freedom, approaching or even exceeding the efficiencies of infinitely flexible swimmers and those with higher degrees of freedom.

\section{Continuum Kinematics}

In this section, we derive the kinematics of the soft continuum swimmer using Lie group theory. We begin by assuming that the deformation of the continuum satisfies the area conservation condition. Under this assumption, the deformation is entirely determined by the change in curvature along the backbone \cite{olson2020euler}. At the normalized time $\vartime$, the transformation from the tail to the frame at the normalized arclength $\arclen$ is given by
\beq
    \locfiber\locfiberpar = \Prodi^{\arclen}_{0}\left(\matrixid + \backboneflowcirc\locfiberparaltarclen d\altarclen\right),
\eeq
where $\backboneflowcirc \in se(2)$ represents the backbone flow along the arclength,\footnote{The ``open circle'' notation used here is similar to the ``dot'' notation for time derivatives but denotes velocity in a local frame rather than in a global coordinate frame.} which can be expressed as
\beq
    \backboneflowcirc\locfiberpar = 
    \transpose{\begin{bmatrix}
        \totalarclen & 0 & \curv\locfiberpar
    \end{bmatrix}},
    \label{eq:backboneflow}
\eeq
where $\totalarclen$ is the total arclength and $\curv\locfiberpar$ is the curvature along the backbone.\footnote{Note that our approach can be generalized to three-dimensional systems or systems with variable total backbone length, area, or shear by introducing additional parameterized variables to the backbone flow in~\eqref{eq:backboneflow}.}

Because the deformation of the continuum---and the resulting kinematics and locomotion---is entirely governed by changes in curvature, which is in turn determined by the design and control of the soft actuators, curvature parameterization is crucial for modeling soft continuum swimmers. In the next subsection, we discuss two different curvature parameterizations: one for soft continuum swimmers defined by the shape modes of the actuators, and another for swimmers with infinite flexibility.

\subsection{Curvature Parameterization}

\begin{figure}[!t]
\centering
\includegraphics[width=\linewidth]{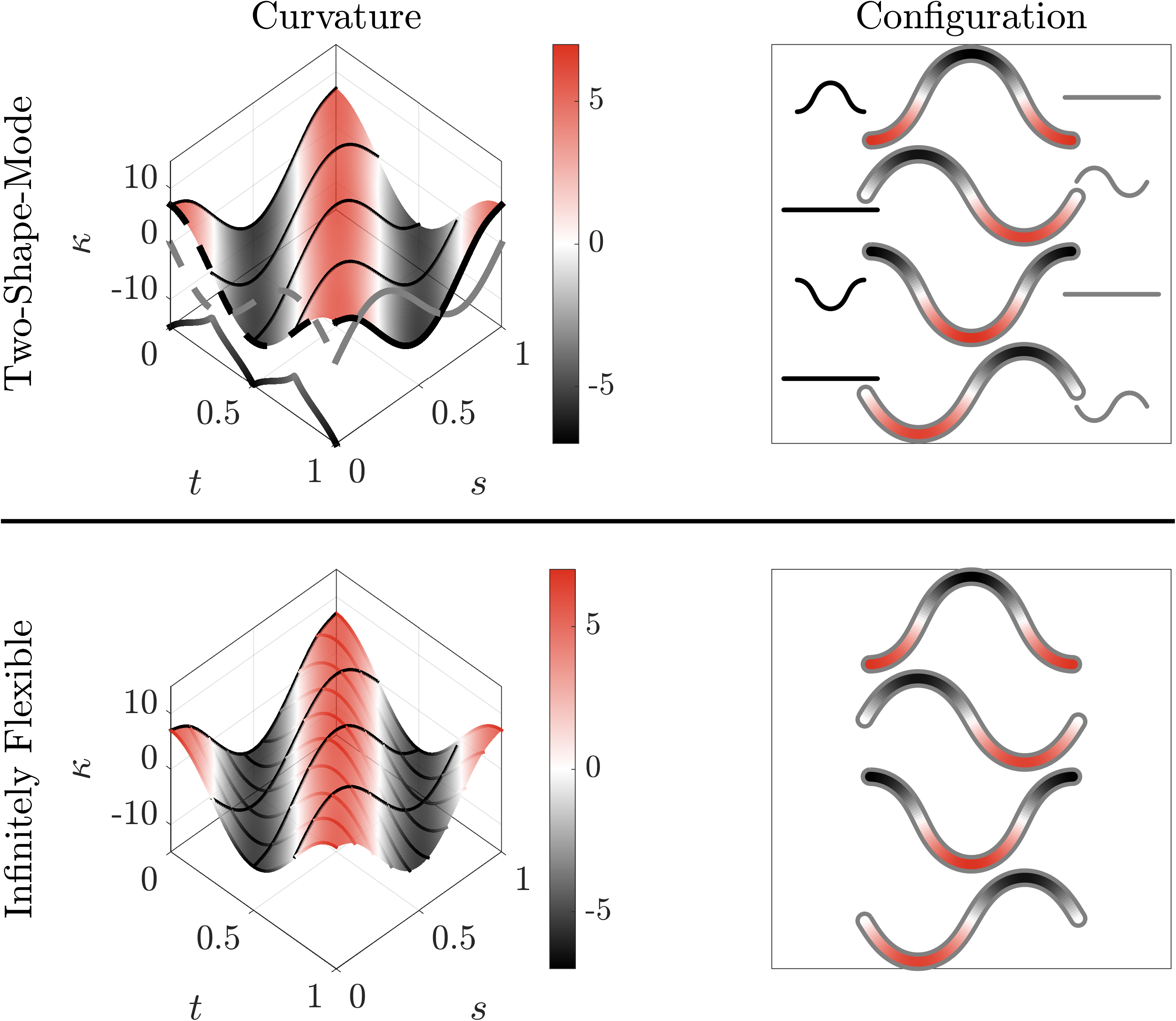}
\caption{Demonstration of the deformation of soft continuum robots parameterized by curvature functions. (Top left): The curvature of a two-shape-mode swimmer, where the curvature is a sum of two shape modes. The shape modes are visualized by black and grey curves at $\vartime = 1$, with shape variables represented by dashed lines at $\arclen = 0$. (Top right): Snapshots of the two-shape-mode swimmer’s configuration corresponding to the black curves in the top left subfigure, with curvature indicated by a red-white-black color scheme. Each shape mode is also visualized as a solid black and grey swimmer on the sides. (Bottom left): The curvature of an infinitely flexible swimmer, parameterized by unit shape modes distributed along the arclength. The unit shape modes, scaled by the shape variables, are visualized by the red-white-black curves. (Bottom right): Snapshots of the infinitely flexible swimmer’s configuration corresponding to the black curves in the bottom left subfigure.}
\label{fig:kinematics}
\end{figure}

The curvature along the backbone is defined as a scalar field, $\curv\locfiberpar$, that maps the normalized arclength, $\arclen$, and time, $\vartime$, to the body curvature \cite{hatton2010generating}. 
We parameterize the curvature as a linear combination of $\fstnum$ shape modes. Shape modes represent deformation patterns encoded by soft actuators, artificial muscles, or, in extreme cases, discrete joints,\footnote{For reference, both the three-link and serpenoid swimmers have two shape modes. The three-link swimmer's shape modes are unit curvatures at two joint locations, while the serpenoid swimmer's shape modes are sine and cosine waves along the backbone.} with deformation amplitudes determined by their corresponding shape variables. The curvature along the body can be generated from the linear combination of these local modes as
\beq
    \curv\locfiberpar = \sum^{\fstnum}_{\fstidx}\curv_{\fstidx}(\arclen)\base_{\fstidx}(\vartime),
\eeq
where $\curv_{\fstidx}(\arclen)$ represents the shape mode that maps arclength to curvature in the mode, and $\base_{\fstidx}(\vartime)$ represents the shape variable that maps time to deformation amplitudes. 

The shape modes and shape variables we consider in this paper are parameterized by third-order polynomials, defined by their control points, as well as continuity and boundary conditions. By stacking the shape modes and shape variables as vectors, the curvature can be expressed as:
\beq
    \curv\locfiberpar = \polybasis_{\curv}\inv{\polyconst}_{\curv}
    \transpose{\begin{bmatrix}
        \curv_c & \curv_c & \boldsymbol{0}
    \end{bmatrix}}
    \begin{bmatrix}
        \base_c & \base_c & \boldsymbol{0}
    \end{bmatrix}\transpose{\left(\inv{\polyconst}_{\base}\right)}\transpose{\polybasis}_{\base},
    \label{eq:par}
\eeq
where $\base_c$ and $\curv_c$ represent the control points defining the polynomials, $\polybasis_{\curv}(\arclen)$ and $\polybasis_{\base}(\vartime)$ are selection matrices used to locate the specific polynomial segment given arclength and time, and $\polyconst_{\base}$ and $\polyconst_{\curv}$ encode the continuity and boundary conditions for the polynomials. Specifically, the continuity conditions ensure that adjacent polynomials at the control points produce consistent values with smooth first- and second-order derivatives, while the boundary conditions require that the curvature acceleration of each shape mode is zero at the head and tail, and that the first- and second-order time derivatives of the shape variables are smooth between the end of one gait cycle and the beginning of the next.

In this paper, we consider two different types of soft continuum swimmers, as illustrated in Fig.~\ref{fig:kinematics}. First, we assume the continuum is infinitely flexible to explore the most efficient deformation patterns for swimming. Next, we address a more realistic scenario where the continuum deformation is controlled by a finite number of shape modes, corresponding to soft actuators or artificial muscles that encode deformation patterns, and seek the co-optimization of the design and control of these shape modes.

For an infinitely flexible continuum, the curvature parameterization can be considered as $\fstnum \to +\infty$ shape modes along the arclength. Linear combinations of these infinite shape modes can generate any possible curvature function. In this case, the control points of the shape modes are unit constants:
\beq
    \curv_{c} = \diag(\underbrace{\begin{bmatrix}
        1 & 1 & \cdots & 1
    \end{bmatrix}}_{\fstnum}).
\eeq
We can then optimize the control points of the shape variables $\base_{c}$ to maximize swimming efficiency.

In the more realistic scenario, we assume that the soft continuum swimmer is driven by a finite number of actuators or artificial muscles. The shape modes correspond to the deformation patterns encoded by these actuators or muscles, and the shape variables represent the actuation amplitudes. In this case, the control points of both the shape modes and shape variables need to be determined, leading to a combined design-and-control optimization problem for the soft continuum swimmer.

\subsection{Continuum Differential Kinematics}
\label{sec:diffkinematics}

The time derivative of the backbone transformation is
\beq
    \tfrac{d}{d\vartime}\locfiber\locfiberpar = \int^{\arclen}_{0}\left(\Adj_{\locfiber\locfiberparaltarclen}\tfrac{d}{d\vartime}\left(\backboneflowcirc\locfiberparaltarclen\right)\locfiber\locfiberpar\right)d\altarclen,
\eeq
where $\Adj_{\bullet}$ is the adjoint operator that converts the group velocity between different frames.\footnote{Formally, $\Adj_{\bullet}$ and $\Adj^{*}_{\bullet}$ refer to the adjoint and dual adjoint operators, respectively. These operators combine the cross-product operation---which converts linear and angular velocities or forces/momenta between different frames---with the rotation operation that expresses velocities or forces/momenta in body-aligned coordinates.} The time derivative of the backbone flow is defined as
\beq
    \tfrac{d}{d\vartime}\backboneflowcirc\locfiberpar = 
    \transpose{\begin{bmatrix}
        0 & 0 & \Curv_{\vartime}\locfiberpar
    \end{bmatrix}},
\eeq
where $\Curv_{\vartime}$ is the time derivative of the curvature. The group velocity of the backbone transformation is 
\begin{eqalign}
    \locfibercirc\locfiberpar &= \inv{\locfiber}\locfiberpar\tfrac{d}{d\vartime}\left(\locfiber\locfiberpar\right) \\
    &= \Adj_{\inv{\locfiber}\locfiberpar}\int^{\arclen}_{0}\left(\Adj_{\locfiber\locfiberparaltarclen}\tfrac{d}{d \vartime}\left(\backboneflowcirc\locfiberparaltarclen\right)\right)d\altarclen.
    \label{eq:localbodyvel}
\end{eqalign}

Considering the floating base position $\fiber$ at the tail,\footnote{In our implementation, we use the geometric center as the floating base for better accuracy in the displacement and gradient approximations discussed later \cite{hatton2011geometrica, bass2022characterizing}. Therefore, the calculation requires an additional adjoint operation and the inclusion of the body velocity of the transformation to the geometric center.} the body velocity of this frame relative to the world is
\beq
    \backbonebodyvel\locfiberpar &= \Adj_{\inv{\locfiber}\locfiberpar}\bodyvel(\vartime) + \locfibercirc\locfiberpar,
    \label{eq:bodyvel}
\eeq
where $\bodyvel$ is the body velocity of the floating base.

\section{Continuum Swimming Dynamics}

\begin{figure}[!t]
\centering
\includegraphics[width=\linewidth]{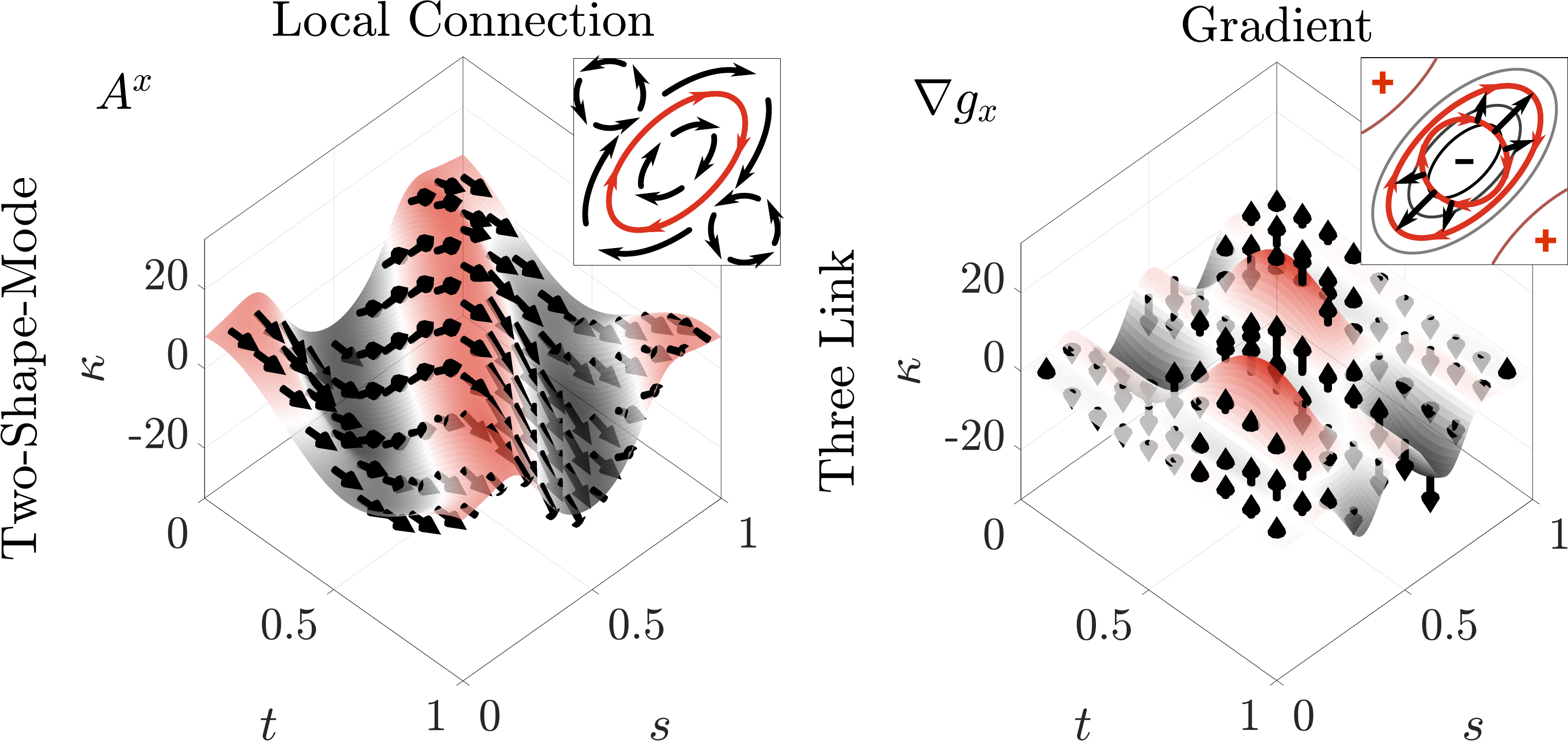}
\caption{Geometric mechanics-based locomotion analysis of soft continuum swimmers. (Left): The local connections in the $x$ direction for a two-shape-mode high-Re swimmer with an optimal design and gait, represented by arrows at each time and arclength. The magnitude of the local connection at each time is scaled to match the tangent vector of the curvature surface in the time direction. (Right): The gradients of displacement in the $x$ direction relative to the magnitude of curvature at each arclength and time for a three-link high-Re swimmer with an optimal gait, represented by arrows. The magnitude of each arrow indicates the increase in displacement for a unit increase in curvature at the arrow's position. The two insets illustrate the concepts of local connection and gradient in gait space.}
\label{fig:local_connection}
\end{figure}

In this section, we discuss the dynamics of soft continuum robots swimming at low and high Re. Both cases can be analyzed using our previous work on the unified geometric representation of locomotion via local metrics---when controlling the system's deformation driven by the curvature changes $\Curv_{\vartime}$, the system's body velocity $\bodyvel$ follows the minimum energy curve defined by the local metrics \cite{yang2024geometric}:
\begin{eqalign}
    \bodyvel(\vartime) &= \mathrm{argmin} \int_{0}^{1}\left(\left\langle\backbonebodyvel\locfiberpar, \backbonebodyvel\locfiberpar\right\rangle_{\localmetrics}\right)d\arclen \label{eq:geneoma} \\
    &= \mathrm{argmin}
    \int^{1}_{0}\left(\left\langle \begin{bmatrix}
    \bodyvel(\vartime) \\
    \Curv_{\vartime}\locfiberpar
    \end{bmatrix},
    \begin{bmatrix}
    \bodyvel(\vartime) \\
    \Curv_{\vartime}\locfiberpar
    \end{bmatrix}
    \right\rangle_{\Localmetrics\locfiberpar}\right)d\arclen,
    \label{eq:geneomb}
\end{eqalign}
where $\langle\bullet, \bullet\rangle$ denotes the inner product, and $\localmetrics$ represents the local metrics, which can be pulled back into the generalized coordinates, becoming $\Localmetrics$. 

A key aspect of the unified geometric representation of locomotion via local metrics \cite{yang2024geometric} is that the dynamics of swimming at both low and high Re share the same format, differing only in the local metrics used \cite{hatton2013geometric}. At very low Re, viscous drag dominates the hydrodynamics, and any inertial effects are immediately damped out. This dominance results in the viscous drag on an infinitesimal segment of the continuum being a linear function of its velocity \cite{cox1970motion}, and the net drag forces and moments experienced by an isolated continuum swimmer interacting with the surrounding fluid being zero. Together, these two consequences form the equation of motion in~\eqref{eq:geneoma} with the local metrics as the drag coefficient matrix:
\beq
    \localmetrics = \diag\left(\begin{bmatrix}
        1 & \dragcoeff & 0
    \end{bmatrix}\right),
\eeq
where $\dragcoeff$ is the viscous drag ratio. 

At very high Re, inertial effects dominate and viscous drag becomes negligible. In this case, because there are no external forces or torques acting on the combined robot-fluid system, the equations of motion can be derived from the conservation of momentum of the combined robot-fluid system, with the local metrics representing the fluid-added inertia. By considering the soft continuum swimmer as a slender body that is neutrally buoyant, with a circular, constant cross-section along its backbone, we can use Lighthill’s slender body theory to compute the fluid-added inertia matrix of each infinitesimal segment of the continuum \cite{lighthill1971largeamplitude, newman1977marine, boyer2008fast, porez2014improved}:
\beq
    \localmetrics = \diag\left(\begin{bmatrix}
        \rho\pi\crossrad^2 & \rho\pi\crossrad^2 & \tfrac{1}{4}\rho\pi\crossrad^4
    \end{bmatrix} + \begin{bmatrix}
        0 & \rho\pi\crossrad^2 & 0
    \end{bmatrix}\right),
\eeq
where $\rho$ is the fluid density and $\crossrad$ is the swimmer's cross-sectional radius. The first term represents the inertia and the second term corresponds to the fluid-added mass of the circular cross-section. This formulation aligns with Lighthill’s theory, which relates reactive forces to the added mass of fluid that acquires momentum through shape deformations.

The pulled-back metrics, $\Localmetrics$, which encode fluid drag dissipation at low Re or fluid-added inertia at high Re, can be partitioned as
\beq
    \Localmetrics = \begin{bmatrix}
        \Localmetrics_{\fiber\fiber} & \Localmetrics_{\fiber\curv} \\ \Localmetrics_{\curv\fiber} & \Localmetrics_{\curv\curv}
    \end{bmatrix}.
\eeq
By substituting $\bodyvel$ with~\eqref{eq:bodyvel} in~\eqref{eq:geneomb} and comparing it to~\eqref{eq:geneoma}, we can derive the upper part of the pulled-back metrics
\begin{eqalign}
    \Localmetrics_{\fiber\fiber}\locfiberpar &= \coAdj_{\inv{\locfiber}\locfiberpar}\localmetrics\Adj_{\inv{\locfiber}\locfiberpar} \\
    \Localmetrics_{\fiber\curv}\locfiberpar &= \left(\int_{\arclen}^{1}\coAdj_{\inv{\locfiber}\locfiberparaltarclen}\localmetrics\Adj_{\inv{\locfiber}\locfiberparaltarclen}d\altarclen\right) \nonumber \\
    &\quad \Adj_{\locfiber\locfiberpar}
    \transpose{\begin{bmatrix}
        0 & 0 & 1
    \end{bmatrix}}.
\end{eqalign}
Note that the minimization in~\eqref{eq:geneomb} depends only on the upper half of $\Localmetrics$, so the body velocity can be solved as
\begin{eqalign}
    \bodyvel(\vartime) &= \int_{0}^{1}\mixedconn\locfiberpar\Curv_{\vartime}\locfiberpar d\arclen \\
    \mixedconn\locfiberpar &= -\inv{\left(\int_{0}^{1}\locmetric_{\fiber\fiber}\locfiberparaltarclen d\altarclen\right)}\locmetric_{\fiber\curv}\locfiberpar,
    \label{eq:bodyvelsol}
\end{eqalign}
where $\mixedconn$ is the local connection describing the relationship between the continuum's deformation and its locomotion. The left panel of Fig.~\ref{fig:local_connection} shows the local connection in the $x$ direction of the optimal design and gait for a two-shape-mode swimmer operating at high Re. Visually, when the changes in body curvature over time align with the local connection arrow at that location and time, the continuum's deformation will propel the swimmer forward; if anti-aligned, it will drive the swimmer in the opposite direction. Better alignment results in greater body velocity.

\section{Geometric Design and Gait Co-Optimization for Soft Continuum Swimming}

As shown in the previous section, the locomotion of a soft continuum swimming robot is determined by its deformation. In this section, we discuss how to effectively optimize this deformation for maximum efficiency using geometric mechanics tools, thereby enabling geometric design and gait co-optimization for soft continuum swimming to maximize efficiency, defined as the displacement divided by power consumption during a cycle:
\beq
    \max_{\text{Deformation}} \left(\tfrac{\text{Displacement}}{\text{Power Consumption}}\right).
\eeq

\subsection{Gradient of Displacement Due to Periodic Deformation}

\begin{figure}[!t]
\centering
\includegraphics[width=0.85\linewidth]{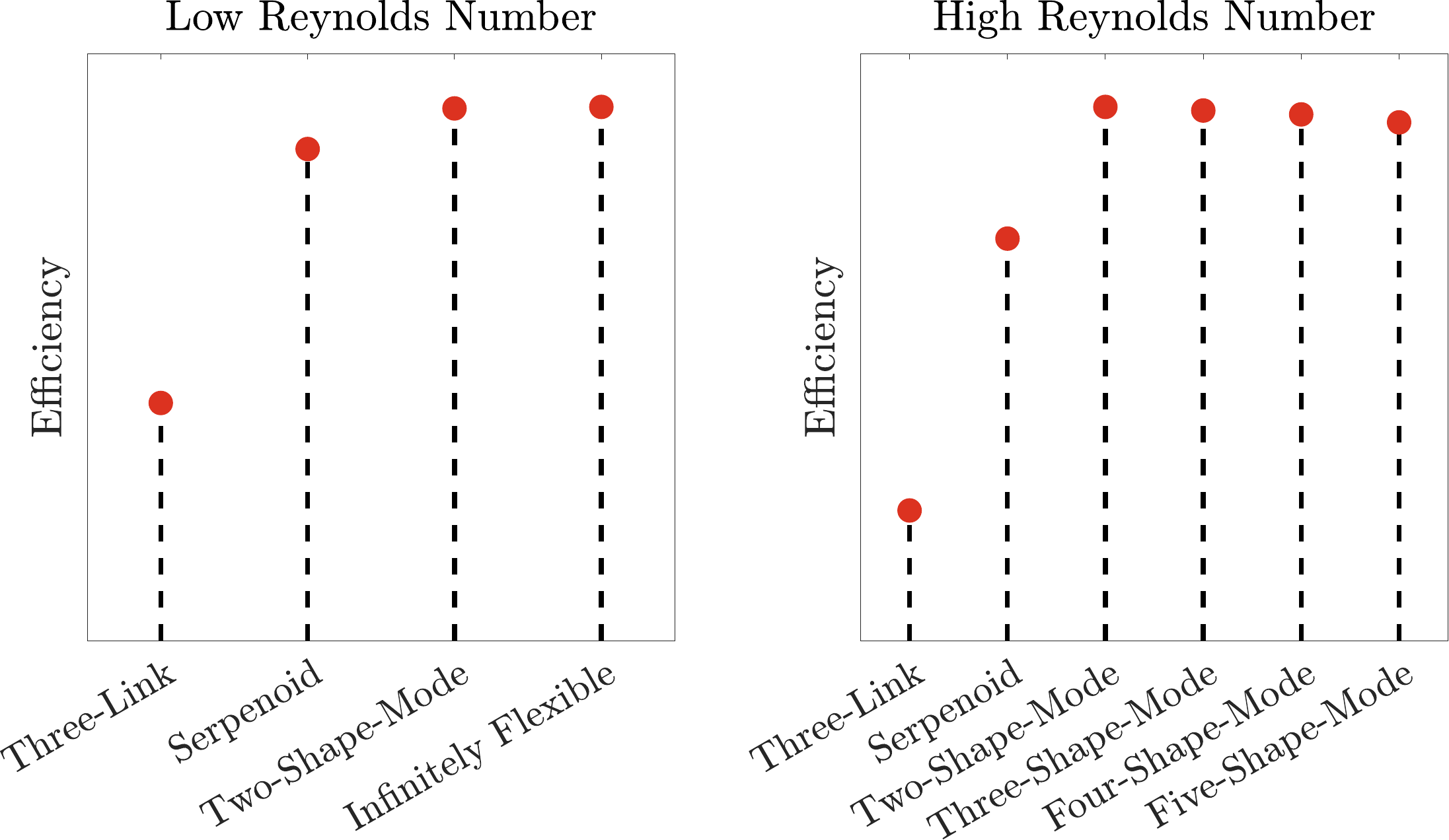}
\caption{Comparison of the optimal design-and-gait efficiencies of different continuum swimmers swimming forward at low and high Re. In both plots, efficiency is normalized by the maximum efficiency. The low Re swimmers use dissipated power metrics based on the Riemannian metrics of the system's shape velocity, while the high Re swimmer uses the covariant acceleration metric related to the square of the active force.}
\label{fig:efficiency}
\end{figure}

Because the drag force or fluid-added momentum is not isotropic in different directions, swimming by cyclic and non-harmonic deformation produces net displacements. Extensive research in the geometric mechanics community \cite{murray1993nonholonomic, walsh1995reorienting, ostrowski1998geometric, melli2006motion, morgansen2007geometric, shammas2007geometric, avron2008geometric, hatton2013geometric, hatton2015nonconservativity, ramasamy2019geometry, hatton2022geometry} has studied the system's \textit{constraint curvature}---a measure of how ``noncanceling'' the system's dynamics are during periodic deformation---to understand which gaits produce useful displacements. Here, we generalize this concept to continuum swimming.

As discussed in \S~\ref{sec:diffkinematics}, the swimmer's body velocity can be determined by applying the local connection $\mixedconn$ to the time derivative of the curvature and then integrating the contributions at each point along the backbone. Consequently, the displacement resulting from a periodic deformation, $\gait$, can be expressed as the line integral of the local connection along the shape change over the deformation period. Building on previous work \cite{radford1998local, melli2006motion, shammas2007geometric, hatton2013geometric, hatton2015nonconservativity}, the exponential coordinates of this displacement, $\log\left(\gaitdisp\right)$, can be approximated by the surface integral of the total Lie bracket $D(\mixedconn)$ of the local connections:
\begin{eqalign}
    \gaitdisp &= \oint_{\gait}\fiber(\vartime)\bodyvel(\vartime)d\vartime \\
    &= \oint_{\gait}\left(\int^{1}_0\fiber(\vartime)\mixedconn\locfiberpar\Curv_{\vartime}\locfiberpar d\arclen\right)d\vartime \\
    \log(\gaitdisp) &\approx \int^1_0\int^1_\arclen\left(\oiint_{\gait} D(\mixedconn)_{\arclen, \altarclen}\right)d\altarclen d\arclen,
\end{eqalign}
where
\beq
D(\mixedconn)_{\arclen, \altarclen} = \left(\der\mixedconn_{\arclen, \altarclen} + [\mixedconn_\arclen, \mixedconn_\altarclen]\right)d\curv(\arclen)\wedge d\curv(\altarclen),
\eeq
in which the differential two-forms $d\curv(\arclen)\wedge d\curv(\altarclen)$ represent the pairs of body curvature changes at arclength $\arclen$ and $\altarclen$. The terms $\der\mixedconn_{\arclen, \altarclen}(\vartime)$ and $[\mixedconn_\arclen, \mixedconn_\altarclen]$ represent the exterior derivative and local Lie bracket, respectively, between $\mixedconn\locfiberpar$ and $\mixedconn\locfiberparaltarclen$. These two terms respectively describe how changes in the local connection due to variations in curvature prevent the net motion from being canceled out over a cycle, and how the net displacement results from the ``parallel parking'' coupling effect between translation and rotation.

Given this displacement approximation, the gradient of the displacement with respect to changes in the deformation pattern can be calculated as the variation of the gait curve, multiplied by the magnitude of the integrand evaluated on the gait. This process can be computed using the generalized Green's theorem:
\beq
\begin{aligned}
    &\nabla\log(\gaitdisp) \approx \\
    &\quad \int^1_0\int^1_0\left(\oint_{\gait} \nabla\curv\locfiberpar\Curv_\vartime\locfiberparaltarclen D(\mixedconn)_{\arclen, \altarclen}d\vartime\right)d\altarclen d\arclen.
\end{aligned}
\eeq
The right panel of Fig.~\ref{fig:local_connection} shows the gradient of displacement in the $x$ direction with respect to the deformation for the optimal gait of a three-link swimmer operating at high Re. Visually, the magnitude of each gradient arrow represents the increase in displacement for each unit increase in body curvature at the arrow's position. The local gradient in the figure suggests that the displacement produced by the gait would be improved by raising the central trench between the two bumps on the body curvature surface to form a ridge, while lowering both sides to form trenches. Optimizing the deformation based on these gradients will result in a pattern similar to the optimal design and gait of the two-shape-mode swimmer shown in the left panel of Fig.~\ref{fig:local_connection}.

Because the deformation is defined by shape modes and shape variables, and ultimately parameterized by their corresponding control points $[\curv_c, \base_c]$ as in~\eqref{eq:par}, with the Jacobian matrix $\jac\locfiberpar = \parderiv{\kappa\locfiberpar}{[\curv_c, \base_c]}$, the displacement gradient can be propagated to the control points. With this gradient formulation, geometric design and gait co-optimization can be efficiently performed using any gradient-based optimizer.

\subsection{Continuum Swimming Power Consumption}

One final consideration in the optimization is the incorporation of appropriate power consumption regularization. In our previous works \cite{ramasamy2019geometry, hatton2022geometry}, we proposed practical power consumption metrics for the different dynamics of swimming at low and high Re. In this paper, we select appropriate metrics to fairly compare swimmers with different shape modes and degrees of freedom at different Re. For low Re, we define the cost of motion as the drag power dissipated into the fluid \cite{ramasamy2019geometry}. For high Re, we define the cost of motion as the covariant acceleration of the body particles, i.e., the acceleration required to follow the gait that is not provided by the natural dynamics of the system \cite{hatton2022geometry}.

\section{Numerical Results}

This section presents simulation results for the optimal design and gait of soft continuum swimmers swimming forward at low and high Re using our framework.\footnote{Please refer to the supplementary video for more details, including demonstrations of swimming in the $y$ and $\theta$ directions.} We compare our optimal solutions with three-link and serpenoid swimmers to demonstrate efficiency improvements. Furthermore, we show that our solution matches or exceeds the efficiency of infinitely flexible swimmers at low Re and swimmers with higher degrees of freedom at high Re.

In our implementation, we discretize the continuum body into 100 segments and use trapezoidal integration for numerical analysis. Each shape mode and shape variable is defined by 10 control points. The infinitely flexible swimmer is parameterized by 10 unit shape modes. At high Re, the swimmer has a slenderness ratio of 100:1, and at low Re, the drag coefficient ratio is 2:1. Simulations are performed using the \texttt{ODE45} function in MATLAB. Deformation optimization is carried out with the \texttt{fmincon} function in MATLAB, with the gradient input calculated using our method. For three-link and snake cases, the initial guess was set as the central circle in the shape space, and these solutions served as the initial guess for higher-order optimization. Self-collision was avoided through constraints. Because the method is gradient-based, the initial guess influences the converged local minimum; assuming a single traveling wave typically leads to a single wave solution. As \cite{wu1961swimming} suggests, exploring multiple traveling waves may enhance efficiency---a direction for future work. The convergence criteria were based on optimality and changes in decision variables, with optimization typically completing in under 5 minutes.

\begin{figure}[!t]
\centering
\includegraphics[width=\linewidth]{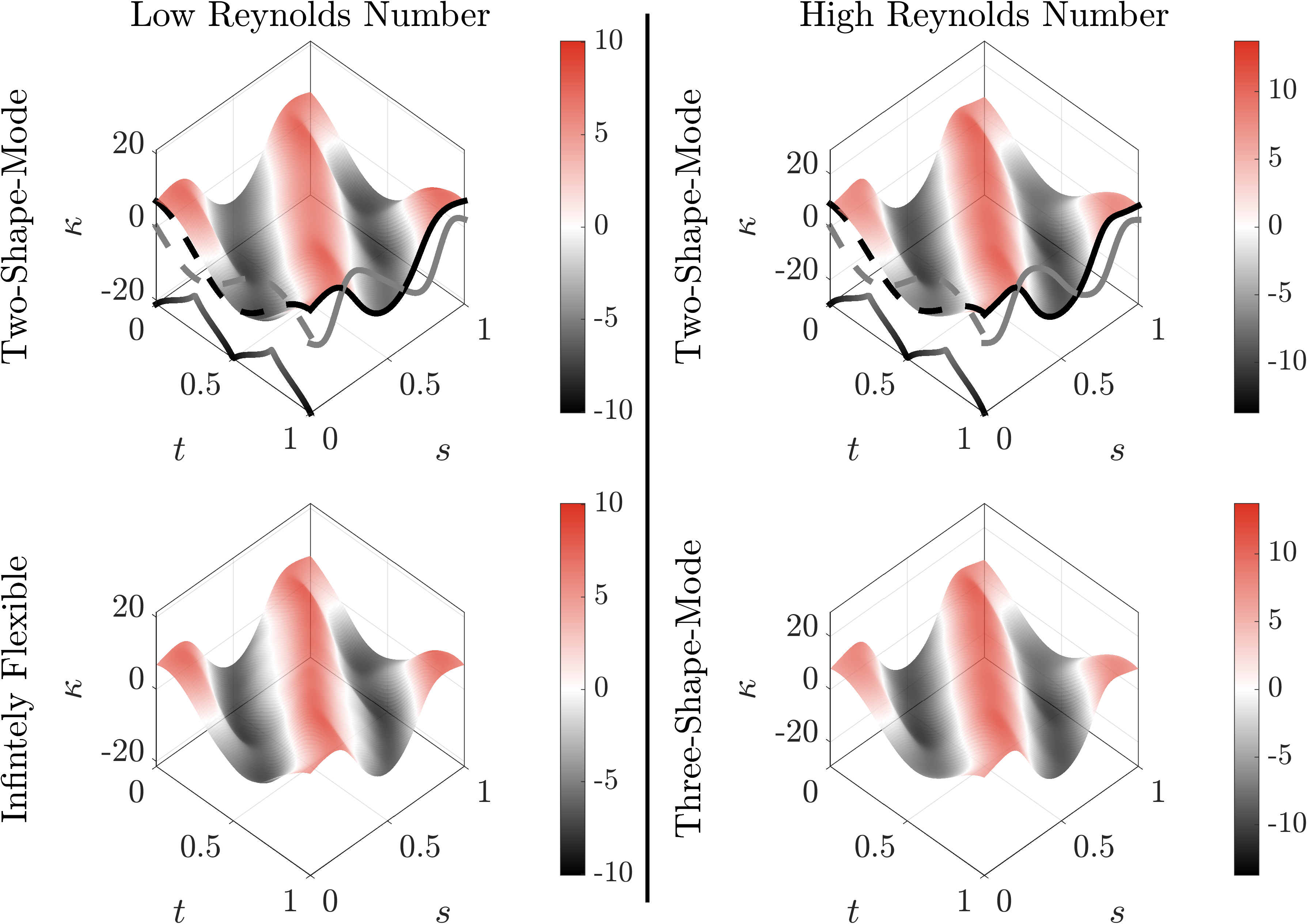}
\caption{Comparison of the optimal forward swimming deformation for determined by the design and gait of the two-shape-mode swimmer with other swimmers at low and high Re. (Left): Comparison of the curvature between the optimal design and gait of the two-shape-mode swimmer and the optimal gait of the infinitely flexible swimmer at low Re. (Right): Comparison of the curvature between the optimal design and gait of the two-shape-mode swimmer and the three-shape-mode swimmer at high Re.}
\label{fig:flexibility_comparison}
\end{figure}

Fig.~\ref{fig:efficiency} illustrates the optimal design-and-gait efficiencies of different continuum swimmers at low and high Re. Overall, the three-link swimmer exhibits lower efficiency, while the two-shape-mode swimmer with optimal design-and-gait demonstrates high efficiency, even when compared to systems with higher degrees of freedom. For swimming at low Re, we consider the dissipated power metrics based on the Riemannian metrics of the system's shape velocity. In this scenario, the serpenoid swimmer performs similarly to the optimal two-shape-mode swimmer. Notably, the optimal design of the two-shape-mode swimmer achieves an efficiency nearly equivalent to that of an infinitely flexible swimmer. This similar efficiency indicates that two shape modes encoding appropriate traveling waves, combined with effective gait modulation, are sufficient to convert all joint actuation power into propulsion without incurring additional dissipation from internal cancellation. This result highlights the utility of considering design-and-gait co-optimization---only two actuators with the optimal design-and-gait are needed to achieve the same efficiency as a swimmer with infinite degrees of freedom. 

For swimming at high Re, we consider the covariant acceleration metric related to the square of the active force applied to the system, which accounts for the dynamics. In this case, the optimal design-and-gait of the two-shape-mode swimmer shows significant efficiency improvements over the serpenoid swimmer, while adding more shape modes results in similar or decreased efficiency due to the additional cost of the extra degrees of freedom.\footnote{Note that the gait cycle is ultimately a circle immersed in shape space, and circles can inherently be described by two phase variables. More than two shape modes serves to handle cases where the cycle is potato-chip-shaped in the infinite-mode shape space.} The efficiency peak at the two-shape-mode again demonstrates the practicality of our approach---design-and-gait co-optimization can achieve a good balance between the number of actuators and efficiency.

Fig.~\ref{fig:flexibility_comparison} compares the optimal design and gait of the two-shape-mode swimmer with other swimmers at low and high Re, respectively. For swimming at low Re, the comparison of curvature between the two-shape-mode swimmer and the infinitely flexible swimmer shows that the two-shape-mode co-optimization successfully reproduces most of the curvature trends of the infinitely flexible swimmer's optimal gait, with the exception of subtle details at the head and tail that cannot be captured by the weighted sum of the two shape modes. The normalized mean squared error between the two-shape-mode swimmer's curvature and that of the infinitely flexible swimmer is approximately 1.3\%. This results in nearly identical efficiency between them, highlighting the practicality of design and gait co-optimization for hardware implementation for soft continuum swimmers. 

For swimming at high Re, the comparison of curvature between the two-shape-mode swimmer and the three-shape-mode swimmer shows that adding the extra shape mode does not significantly alter the curvature. In fact, to avoid additional power consumption due to the extra degrees of freedom, the optimization tends to make the additional shape mode of the three-shape-mode swimmer as close as possible to being orthogonal to the original motion, thereby minimizing power consumption. 

\begin{figure}[!t]
\centering
\includegraphics[width=\linewidth]{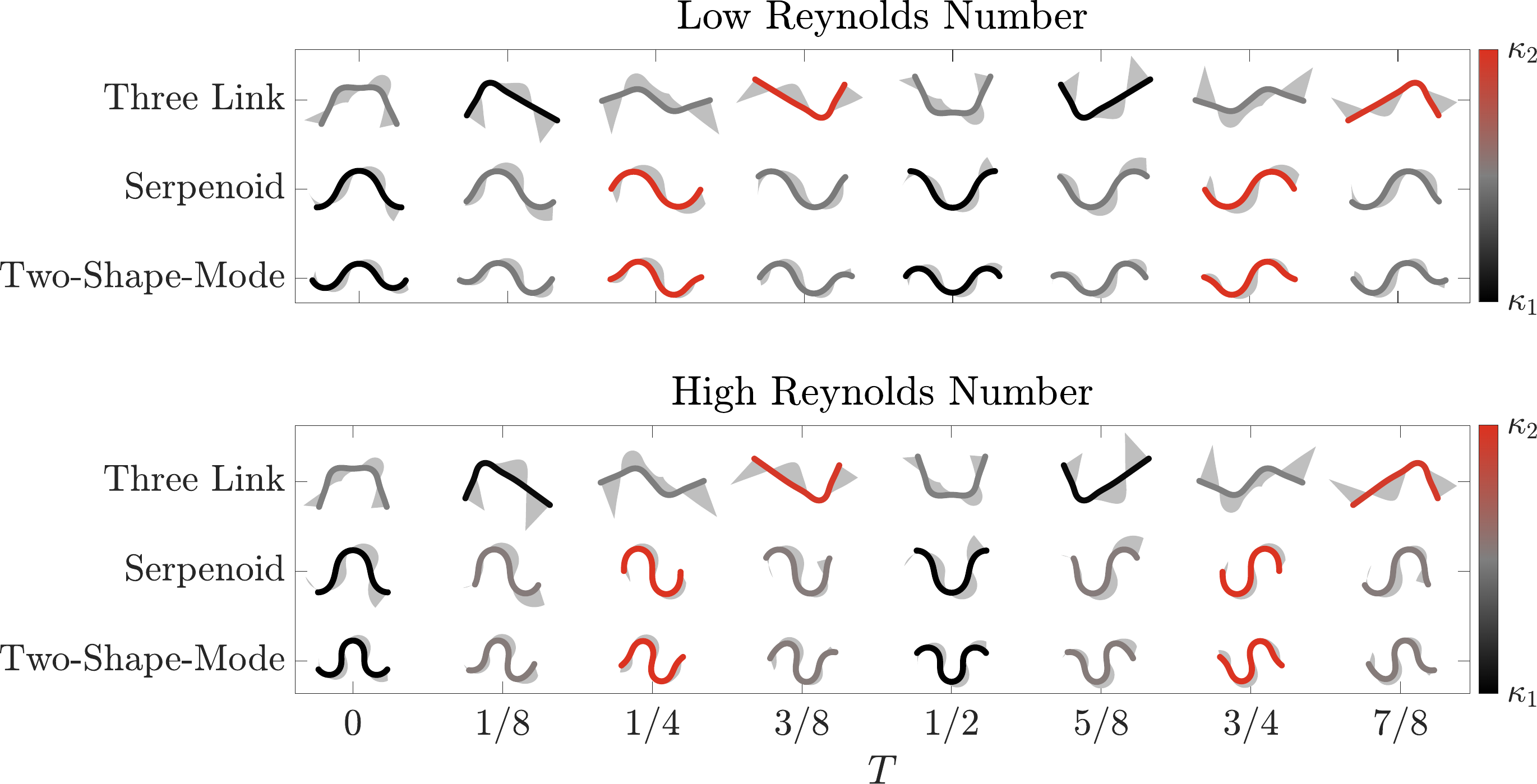}
\caption{Snapshots of the optimal shape mode design and gaits for soft continuum swimmers swimming forward at low and high Re at different times in the period $T$. The color of the swimmer indicates the ratio of the two shape modes dominating the deformation at each moment. The gray fins represent the drag force or negative linear momentum on each segment of the swimmer at that moment.}
\label{fig:gait}
\end{figure}

Fig.~\ref{fig:gait} shows the optimal deformation of different swimmers at low and high Re. For the three-link and serpenoid swimmers, we fixed the shape mode and optimized only the gait, while for the two-shape-mode swimmer, both shape modes and gaits were optimized. At low Re, the deformation becomes progressively smoother and more consistent from the three-link to the two-shape-mode swimmer, leading to a more uniform drag distribution on the continuum and reduced power dissipation due to viscous drag. 

Similarly, for swimming at high Re, we observe that from top to bottom, the sweep of the continuum during locomotion becomes increasingly gentle, resulting in a more consistent momentum distribution across each segment during deformation, thus minimizing the active force required to change momentum during swimming.

Comparing behaviors at low and high Re reveals that high-Re swimmers exhibit greater bending. This difference may arise from the definitions of power consumption: at low Re, power dissipation scales with the square of velocity, whereas at high Re, it is based on the square of covariant acceleration. Although the dynamics at low and high Re are not directly linked, this insight could inform future studies on swimming at intermediate Re.

\section{Conclusion and Future Work}

In this paper, we have presented a practical framework for the co-optimization of the design and control of soft continuum robots, leveraging geometric mechanics-based locomotion analysis for swimming at low and high Re. By extending geometric principles to continuum bodies, our approach enables geometric variational co-optimization of designs and gaits across various power consumption metrics and swimming environments. The resulting optimal solutions exhibit great efficiency at both low and high Re compared to three-link or serpenoid swimmers with the same degrees of freedom, approaching or even exceeding the efficiencies of infinitely flexible swimmers and those with higher degrees of freedom. These promising results highlight the significant potential of our framework to guide the design and control of efficient soft continuum robots.

Future work includes applying our framework to real robots \cite{jusufi2017undulatory, vanstratum2022comparative}, refining soft actuator models to account for response speed, durability, and fatigue during optimization \cite{olson2020euler, godage2024modeling}, and exploring multi-material soft robots \cite{phamduy2017design}. We will also investigate more complex locomotion tasks \cite{hatton2013snakes, choi2022optimal} and explore more complicated hydrodynamics with non-zero momentum effects \cite{shammas2007unified, kelly2012proportional, yang2023geometric}. Additionally, we plan to extend this framework to jet propulsion systems \cite{jones2024underwater} and refine coordinate optimization to improve displacement approximation \cite{yang2024geometric}. Compared to \cite{wu1961swimming}, our solution is consistent in that the motion wave propagates from head to tail. However, our model does not exhibit the linearly increasing amplitude observed in \cite{wu1961swimming}, possibly due to the omission of vortex shedding and second-order dynamics. Incorporating a dynamic model could better capture these effects and extend our approach to scenarios at intermediate Reynolds numbers, where both fluid drag and added mass are significant \cite{porez2014improved}, although this would complicate the optimization due to time-variant dynamics.

\bibliographystyle{IEEEtran}
\bibliography{ref}

\end{document}